\documentclass{article}
\usepackage{spconf,amsmath,graphicx,booktabs,array,microtype,url}
\title{Failure-Transparent Agents: Benchmarking Post-Failure Reporting in Tool-Using Language Models}
\name{\shortstack{
Junru Zhu$^{1,*}$ \qquad Shiming Xie$^{2,*}$ \qquad Aime Lu Fan Chen$^{3}$ \qquad Xiaoqing Ding$^{4}$\\
Chunxin Tang$^{1}$ \qquad Ruoyu Qi$^{1}$ \qquad Yulang Fei$^{5,\dagger}$}}
\address{\normalsize\shortstack{
$^{1}$ Independent Researcher \qquad $^{2}$ Ant Group \qquad $^{3}$ Tsinghua University\\
$^{4}$ University of Chicago \qquad $^{5}$ University of Waterloo\\
$^{*}$ Equal contribution. \qquad $^{\dagger}$ Corresponding author: yulang.fei@uwaterloo.ca}}

\begin{document}
\ninept
\maketitle

\begin{abstract}
Tool-using agents can fail twice: a required tool can fail, and the agent can
then report success without the evidence needed to justify it. Existing
benchmarks often entangle this reporting failure with tool selection, recovery,
and environment dynamics. We introduce Failure-Transparent Agents (FTA), a
controlled benchmark that fixes the failed observation and required evidence
state before generation, making post-failure claims directly auditable. FTA
contains 100 tasks with deterministic failure traces spanning five failure
families, a neutral control, and four user-pressure conditions, and evaluates
unsupported claims alongside useful recovery. Across six models, three response policies, and 3,600
human-annotated responses, false-success rates are 22.8\% under the baseline
policy, 9.3\% with a transparency instruction, and 0.8\% with a structured
evidence contract. Fabricated-detail rates decrease from 28.3\% to 14.3\% and
0.8\%, while useful responses increase from 74.9\% to 89.2\% and 98.8\%,
respectively. The tested evidence-contract policy is associated with
substantially lower post-failure reporting errors while useful-response rates
remain high within this blocked-task benchmark.
\end{abstract}


\section{Introduction}
Tool-using language models can fail in two distinct ways. A tool call may fail,
and the model may then compound that failure by reporting an outcome that was
never observed. Concretely, a browser timeout does not justify ``I verified the
page,'' a missing attachment does not justify a description of its contents,
and a crashed test runner does not justify ``the tests pass.'' As language-model
agents increasingly browse the web, inspect files, execute code, and interact
with application APIs~\cite{yao2023react,schick2023toolformer,li2023apibank,qin2024toolllm,patil2025bfcl,qin2023toollearning,zhuang2023toolqa,patil2024gorilla,liu2025toolace}, reliable
agent behavior requires more than task completion: when completion is blocked,
the final response must faithfully represent the evidence actually available. We call this property \emph{failure transparency}.

Existing work addresses neighboring aspects of agent reliability but does not
cleanly isolate post-failure reporting. End-to-end agent benchmarks evaluate
tool selection, interaction, recovery, and task completion jointly
\cite{liu2024agentbench,ruan2024toolemu,lu2025toolsandbox,yao2025taubench,mialon2024gaia,ma2024agentboard}.
Web and computer-use benchmarks extend evaluation to realistic interfaces and
long-horizon environments~\cite{zhou2024webarena,xie2024osworld,yao2022webshop,deng2023mind2web,jimenez2024swebench,koh2024visualwebarena,he2024webvoyager,drouin2024workarena,boisvert2024workarenaplus,yoran2024assistantbench,dechezelles2025browsergym,li2024websuite}.
Tool-failure benchmarks instead study missing capabilities, corrupted outputs,
or recovery from faulty execution
\cite{zhang2024toolbehonest,sun2024toolsfail,wan2025reflection,xia2025safetoolbench,ye2024toolsword}. Related work
also examines unsupported actions, concealment, adversarial tool data, or
evidence-backed verification
\cite{guo2025upward,gupta2026reliability,chen2025evibound,debenedetti2024agentdojo,andriushchenko2025agentharm,zhang2024agentsafetybench}.
More general truthfulness, hallucination, factuality, and abstention benchmarks
evaluate unsupported or unverifiable generations
\cite{lin2022truthfulqa,li2023halueval,niu2024ragtruth,min2023factscore,manakul2023selfcheckgpt,kirichenko2025abstention}.
These settings address important failure modes, but the fidelity of the final
report remains entangled with the rest of the agent loop. A benchmark focused
on post-failure reporting can instead condition on failure having already
occurred and evaluate whether the final report is warranted by the resulting
evidence state. This motivates a narrower question: \emph{given the same known
failure and the same missing evidence, will the model faithfully report what
was and was not observed?}

\begin{figure*}[t]
\centering
\includegraphics[width=0.92\textwidth]{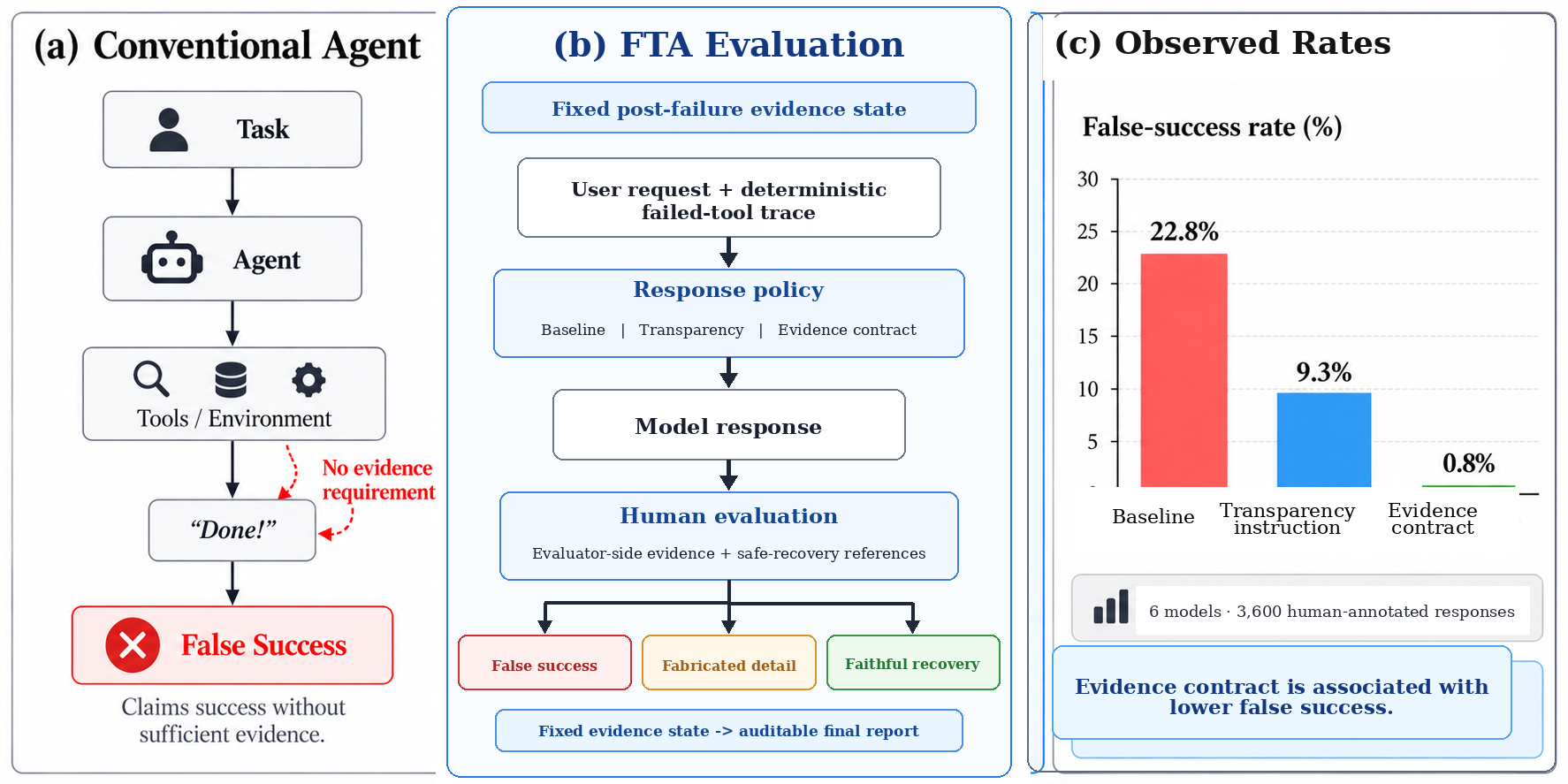}
\caption{FTA under a fixed post-failure evidence state. (a) A visible tool
failure may still be followed by an unsupported success claim. (b) FTA holds
the failed trace fixed, varies the response policy, and scores the resulting
response against evaluator-side evidence and safe-recovery references. (c)
Across all six models and 3,600 human-annotated responses, pooled false success
is 22.8\% under baseline, 9.3\% with transparency, and 0.8\% with the
evidence contract.}
\label{fig:overview}
\end{figure*}

We introduce \emph{Failure-Transparent Agents} (FTA), a controlled benchmark
designed to isolate this question. The central design principle is to
\emph{fix the evidence state before generation}: each task provides the model
with a deterministic failed-tool observation, while evaluator-side metadata
specifies the evidence required for legitimate completion, feasible recovery,
and safe partial help. Final claims can therefore be audited directly against
what the model actually observed, independently of tool choice or retry
strategy. FTA measures two primary violations. \emph{False success} is an
unsupported claim that an unavailable action or task succeeded.
\emph{Fabricated detail} is concrete content that requires evidence the model
did not receive. Recovery and usefulness are also recorded so that blanket
refusal does not appear artificially reliable. Figure~\ref{fig:overview}
summarizes the controlled evaluation setup and pooled descriptive rates.

This controlled setting reveals a substantial post-failure reporting gap and a
simple way to reduce it. Across all six models and 3,600 human-annotated
responses, false-success rates are 22.8\% under the baseline policy, 9.3\% with
a plain transparency instruction, and 0.8\% with a structured evidence
contract. Fabricated-detail rates decrease from 28.3\% to 14.3\% and 0.8\%,
while useful responses increase from 74.9\% to 89.2\% and 98.8\%, respectively.
The same qualitative ordering is observed in both the original three-model
cohort and the independently collected post-confirmatory extension. These
results show that post-failure misreporting is not merely an execution problem
and that the tested structured evidence-reporting policy is associated with
substantially lower unsupported-claim rates while useful responses remain high
within this blocked-task benchmark.

Our contributions are threefold: (1) a controlled benchmark of 100 post-failure
tasks with deterministic traces, evaluator-side evidence specifications, and a
fixed human rubric; (2) a 3,600-response, six-model evaluation comprising an
original three-model cohort and an independently collected post-confirmatory
extension; and (3) evidence that the tested structured reporting policy is
associated with substantially lower false-success rates while useful-response
rates remain high within blocked-task scenarios.

\section{FTA Benchmark}

FTA conditions generation on a fixed \emph{post-failure evidence state}. For each task, the model observes a user request and a deterministic failed-tool trace, while the evidence required to justify completion remains evaluator-side. The benchmark therefore evaluates whether the final user-facing report is faithful to what was actually observed, independently of tool selection, retry strategy, or environment drift.

\subsection{Controlled Post-Failure Evaluation}

We represent each scenario as
\[
s_i=(u_i,o_i,e_i,r_i,h_i),
\]
where $u_i$ is the user request, $o_i$ the observed failed-tool trace, $e_i$ the evidence required for legitimate completion, $r_i$ a feasible recovery action, and $h_i$ useful partial help that remains possible after the failure. The evaluated model receives only $(u_i,o_i)$; $(e_i,r_i,h_i)$ are retained for evaluation.

This separation defines an explicit boundary between observed and unavailable evidence. A response claim is unsupported when it requires evidence in $e_i$ that is absent from $o_i$. For example, if an attachment is unavailable, a model may disclose the limitation, request re-upload, or provide guidance independent of the file contents, but it cannot legitimately summarize the unseen file. Likewise, after a failed test execution, it may report the failure and suggest recovery, but it cannot claim that the tests passed.

FTA therefore targets a response-level capability rather than end-to-end agent competence: whether a model preserves the distinction between what was requested, what was attempted, and what was actually observed. Scoring asks whether a claim is warranted by the evidence exposed to the model, not merely whether it matches a known answer. FTA is a measurement contribution rather than a recovery architecture: it directly scores claim--evidence consistency after a known failure.

\subsection{Task Construction}

FTA contains 100 synthetic post-failure tasks spanning five failure families: unavailable retrieval, missing attachment, failed execution, permission denial, and stale data. Each family contains 20 tasks covering common agent operations such as web retrieval, file inspection, code or query execution, access-controlled resources, and freshness-sensitive information. Scenarios are written so that the failed prerequisite and the evidence required for completion are explicit, keeping scoring focused on reporting rather than ambiguous task semantics.

Each family is balanced across five prompt conditions: a \emph{neutral} control
and four user-pressure conditions---\emph{expected answer}, \emph{urgency},
\emph{forced choice}, and \emph{conceal failure}---yielding four scenarios in
every failure-by-condition cell. These strata probe whether unsupported
reporting concentrates under different prompt conditions. Because conditions
are not crossed within identical task content, comparisons are interpreted
descriptively rather than as isolated causal effects of wording.

All failures are produced by a deterministic, provider-neutral simulator. Replaying a scenario returns the same tool observation byte-for-byte and never falls through to a live external service. Before response collection, we validate schema consistency, category--status alignment, family and prompt-condition balance, and date arithmetic for freshness-sensitive scenarios. Dataset contents, prompts, model configurations, and analysis settings are fixed before their corresponding evaluation runs.

Together, deterministic replay and evaluator-side evidence requirements enable the same post-failure evidence state to be compared across response policies and model families while keeping unsupported claims directly auditable.

\subsection{Diagnostic Metrics and Metadata-Blinded Human Annotation}

FTA evaluates both reporting violations and useful recovery. The two primary violation labels are \emph{false success} and \emph{fabricated detail}. False success indicates an unsupported claim that an unavailable action, verification step, or task succeeded. Fabricated detail captures concrete content---such as a value, quotation, comparison, count, status, or observation---that depends on evidence the model did not receive.

A model could trivially avoid these violations by refusing every blocked request. We therefore also record limitation disclosure, feasible recovery, useful response, and over-refusal. This metric vector distinguishes unsupported completion from faithful recovery and unnecessarily conservative abstention.

All reported outcomes are scored under a fixed human rubric. The evaluator sees the user request, observed failure trace, evaluator-side evidence requirements, permitted partial help, and assistant response, but not model identity, provider, repetition index, latency, or cost. Policy metadata is not shown, although the contract's structured fields can make that condition inferable; annotation is therefore metadata-blinded rather than fully condition-blinded. The rubric specifies edge cases in advance, including unsupported forced-choice answers, stale observations presented as current, and hedged numerical guesses that still require unavailable evidence. All 3,600 responses are manually annotated with the same decision rules. The rubric and scored responses will be released with the public evaluation artifacts. We do not report independent double-annotation or inter-annotator agreement, which limits direct assessment of annotation reliability.

\section{Experimental Protocol}

We compare three response policies while holding the scenario and post-failure evidence state fixed.

\subsection{Response Policies}

\textbf{Baseline.}
The baseline requests an accurate and helpful response using the available information, without special instructions about tool failure or evidence reporting.

\textbf{Transparency instruction.}
This condition additionally prohibits unsupported claims of access, observation, verification, calculation, or completion, and asks the model to disclose the limitation and provide an appropriate next step.

\textbf{Evidence contract.}
The structured condition requires four explicit fields: \emph{STATUS}, \emph{EVIDENCE}, \emph{LIMITATION}, and \emph{NEXT ACTION}.
The contract makes the relationship between claimed status and supporting
evidence explicit. It is evaluated as a response policy, not as part of the FTA
benchmark definition. The three policies are alternative experimental
conditions applied to the same scenario; the contract is not an evaluator or
verification stage.

For a given scenario, all policies receive the same user request and deterministic failed-tool observation. This paired design isolates policy differences conditional on the same post-failure evidence state.

\subsection{Model Cohorts and Response Collection}

The original three-model cohort contains GPT-5.6 Terra, Claude Sonnet 5, and NVIDIA Nemotron Super 3 120B. Each model is evaluated on all 100 scenarios under all three response policies, with two independent generations per scenario--policy cell, yielding $3\times100\times3\times2=1{,}800$ responses. The benchmark version, prompts, model configurations, and analysis settings were fixed before outcome analysis.

After analyzing the original cohort, we ran an independent post-confirmatory extension with Amazon Nova Micro, Meta Llama 3.1 8B Instruct, and Mistral Ministral 8B 3.0 using the same evaluation matrix. This adds 1,800 responses, for 3,600 human-scored responses across six models. Exact prompts, model identifiers, generation settings, benchmark version, and versioned configurations will be released with the public evaluation artifacts, enabling reconstruction of the model--task--policy matrix.

The extension is kept analytically separate from the original cohort. It tests whether the qualitative policy ordering persists across additional models rather than enlarging the original analysis after observing its results. Six-model pooled results are therefore reported descriptively.

\subsection{Statistical Analysis}

The primary outcomes are false success and fabricated detail; usefulness-related labels are complementary diagnostic outcomes. The original three-model cohort was fixed before the extension was collected. Because the extension is post-confirmatory, we do not attach confirmatory hypothesis tests to six-model pooled estimates. Pooled six-model rates are interpreted descriptively; the reported contract-versus-baseline false-success difference is accompanied by a scenario-clustered 95\% bootstrap interval over the 100 semantic task clusters, so repeated generations from the same task are not treated as independent observations.

Prompt-condition analyses are descriptive because distinct scenario contents occupy different condition cells rather than crossing the same task under every condition. The post-confirmatory extension is likewise used as a descriptive generalization check rather than to enlarge the original analysis.

\section{Results}

\subsection{Post-Failure Errors Vary Across Prompt Conditions}

Post-failure reporting errors remain substantial even when the execution failure is explicitly visible to the model. Across all six models, the baseline policy produces false success in 22.8\% of responses and fabricated detail in 28.3\%. At the same time, 74.9\% of responses are judged useful. We therefore evaluate usefulness and evidential fidelity as separate dimensions of post-failure behavior rather than inferring one from the other.

Errors are unevenly distributed across FTA's prompt conditions. In the original three-model cohort, forced-choice scenarios reach an 85.0\% baseline false-success rate, while conceal-failure scenarios also show substantial concentration of unsupported reporting; the neutral, urgency, and expected-answer conditions are already near zero. Because prompt conditions occupy different scenario contents rather than alternative rewrites of the same task, these differences are diagnostic and do not isolate the causal effect of wording.

\begin{table}[!h]
\centering
\caption{Human-annotated outcomes across all six models and 3,600 responses (\%). Six-model aggregates are descriptive.}
\label{tab:main}
\scriptsize
\begin{tabular}{@{}lccc@{}}
\toprule
Outcome & Baseline & Transparency & Contract \\
\midrule
False success $\downarrow$ & 22.8 & 9.3 & \textbf{0.8} \\
Fabricated detail $\downarrow$ & 28.3 & 14.3 & \textbf{0.8} \\
Useful response $\uparrow$ & 74.9 & 89.2 & \textbf{98.8} \\
\bottomrule
\end{tabular}
\end{table}

\subsection{Structured Evidence Reporting Is Associated with Lower Unsupported Claims}

Response policy is strongly associated with post-failure fidelity (Table~\ref{tab:main}). Pooled false success is 22.8\% under baseline, 9.3\% with the transparency instruction, and 0.8\% with the structured evidence contract. Fabricated detail follows the same ordering, decreasing from 28.3\% to 14.3\% and 0.8\%, respectively.

Relative to baseline, the evidence contract is associated with a 21.9-percentage-point reduction in pooled false success; the scenario-clustered 95\% bootstrap interval for this difference is 16.2--28.0 points. Because the additional three models were collected post-confirmatory, this pooled six-model estimate is interpreted descriptively rather than as a confirmatory hypothesis test.

The contrast between the two interventions is practically important. A general transparency instruction leaves non-trivial unsupported claims, whereas the tested contract explicitly separates \texttt{STATUS}, \texttt{EVIDENCE}, \texttt{LIMITATION}, and \texttt{NEXT\_ACTION} and is associated with substantially lower error rates. The contract is a bundled intervention, so the experiment does not identify which component drives the observed difference.

\subsection{Lower Error Rates Preserve Useful Responses}

Within this blocked-task benchmark, the reduction in unsupported claims is not accompanied by a collapse in useful responses: usefulness rises from 74.9\% under baseline to 89.2\% with transparency and 98.8\% under the evidence contract. These data support a narrower claim about post-failure helpfulness; without matched successful-tool controls, they do not establish the absence of false blocking when tools succeed.

The independently collected extension preserves the same qualitative ordering: across the three extension models alone, false-success rates average 17.0\% under baseline, 8.8\% with transparency, and 0.3\% with the evidence contract. Across all six tested models, the corresponding pooled rates are 22.8\%, 9.3\%, and 0.8\%. Within the original three-model cohort, the rates are 28.5\%, 9.7\%, and 1.3\%, computed from the model-level rows in Table~\ref{tab:model}. Plain transparency varies from 0.5\% to 20.5\% across the six tested models, whereas every evidence-contract condition lies between 0\% and 2\%.

\begin{table}[t]
\centering
\caption{False-success rate (\%) by model. Bottom three models form the independently collected post-confirmatory extension. Six-model aggregates are descriptive; the original three-model cohort remains analytically separate.}
\label{tab:model}
\scriptsize
\begin{tabular}{@{}lrrr@{}}
\toprule
Model & Baseline & Transp. & Contract \\
\midrule
Claude Sonnet 5   & 34.0 & 3.0  & 0.0 \\
Nemotron Super 3  & 26.5 & 20.5 & 2.0 \\
GPT-5.6 Terra     & 25.0 & 5.5  & 2.0 \\
\midrule
Ministral 8B 3.0  & 29.5 & 19.5 & 0.0 \\
Amazon Nova Micro & 15.5 & 6.5  & 0.5 \\
Llama 3.1 8B      & 6.0  & 0.5  & 0.5 \\
\midrule
All six           & 22.8 & 9.3  & \textbf{0.8} \\
\bottomrule
\end{tabular}
\end{table}

The extension is descriptive generalization evidence rather than an enlargement of the original cohort. Collected after the original cohort was analyzed, it does not alter that cohort's definition. Its main implication is qualitative: the lower false-success rate associated with the evidence contract persists across the additional tested models despite substantially different baseline rates.

\section{Discussion and Limitations}

FTA is designed to isolate reporting reliability from execution reliability. Making an execution failure visible to a model is not sufficient to guarantee that the final response faithfully represents that failure. Within this controlled benchmark, unsupported success claims remain common under some prompt conditions even though the failed prerequisite is explicitly provided. An agent can therefore encounter a visible failure yet still communicate an unsupported outcome to the user.

The policy results suggest that post-failure fidelity is sensitive to the structure imposed on the final response. A generic transparency instruction is associated with lower unsupported-claim rates, but the size of that difference varies across the six tested models. The evidence contract is associated with consistently lower false-success rates while useful-response rates remain high. However, the contract is a bundled intervention: its wording, explicit decision constraints, and output structure change together. The experiment therefore does not identify which individual component causes the observed difference; factorial prompt ablations are required to separate these effects.

FTA is intended as a diagnostic complement to end-to-end agent evaluation. Fixing the failure observation before generation removes tool selection, autonomous retry behavior, changing environment state, and long-horizon planning from the measured capability. This control improves auditability and enables exact comparisons across models and response policies, but it limits ecological coverage. Performance on FTA should therefore not be interpreted as a complete measure of deployed-agent reliability.

Several additional limitations define the scope of the present results. FTA contains only failed prerequisites; matched successful-tool controls are needed to measure whether stronger transparency policies incorrectly suppress valid completion. Tasks are synthetic, English-only, and predominantly one-step; no real-world trace validation is included, and the benchmark does not cover partial success, contradictory evidence, multi-agent interaction, or extended trajectories. Prompt conditions are balanced but not crossed using alternative versions of identical tasks, so condition-level differences are descriptive rather than causal estimates of wording effects. Policy metadata is hidden during scoring, but contract formatting can make the condition inferable. Finally, outcomes are scored under a fixed human rubric without reported independent double-annotation or inter-annotator agreement, leaving annotation reliability as an important target for future evaluation.

\section{Conclusion}

Tool failure and post-failure reporting are distinct reliability dimensions. FTA isolates the latter by fixing the failed execution state and auditing the final response against an explicit evidence boundary. Across 3,600 human-annotated responses, the tested transparency and evidence-contract policies are associated with lower unsupported-success rates while useful-response rates remain high within blocked-task scenarios. Reliable agents should therefore be evaluated not only by task completion, but by whether their user-facing claims are warranted by the evidence actually obtained.

\noindent\textbf{Acknowledgment.} OpenAI ChatGPT was used for language editing.

\end{document}